\documentclass{article}
\usepackage{arxiv}
\usepackage[utf8]{inputenc} 
\usepackage[T1]{fontenc}    
\usepackage{hyperref}       
\usepackage{url}            
\usepackage{booktabs}       
\usepackage{amsfonts}       
\usepackage{nicefrac}       
\usepackage{microtype}      
\usepackage{lipsum}
\usepackage{graphicx}
\usepackage{color}
\usepackage{soul}
\usepackage{amssymb}
\usepackage{mathtools}
\usepackage{float}
\usepackage{array}
\newcolumntype{C}{>{\centering\arraybackslash}p{4cm}}
\newcolumntype{D}{>{\centering\arraybackslash}p{8cm}}
\usepackage[caption = false]{subfig}

\title{Learning User Preferences via Reinforcement Learning with Spatial Interface Valuing}

\author{
  Miguel ALonso Jr.\thanks{Supported in part by Kynetic AI, LLC \url{https://kynetic.ai}} \\
  School of Computing and Information Science\\
  Florida International University University\\
  Miami, FL 33199 \\
  \texttt{malonsoj@cs.fiu.edu} \\
}

\begin{document}
\maketitle

\begin{abstract}
Interactive Machine Learning is concerned with creating systems that operate in environments alongside humans to achieve a task. A typical use is to extend or amplify the capabilities of a human in cognitive or physical ways, requiring the machine to adapt to the users' intentions and preferences. Often, this takes the form of a human operator providing some type of feedback to the user, which can be explicit feedback, implicit feedback, or a combination of both. Explicit feedback, such as through a mouse click, carries a high cognitive load. The focus of this study is to extend the current state of the art in interactive machine learning by demonstrating that agents can learn a human user's behavior and adapt to preferences with a reduced amount of explicit human feedback in a mixed feedback setting. The learning agent perceives a value of its own behavior from hand gestures given via a spatial interface. This feedback mechanism is termed Spatial Interface Valuing. This method is evaluated experimentally in a simulated environment for a grasping task using a robotic arm with variable grip settings. Preliminary results indicate that learning agents using spatial interface valuing can learn a value function mapping spatial gestures to expected future rewards much more quickly as compared to those same agents just receiving explicit feedback, demonstrating that an agent perceiving feedback from a human user via a spatial interface can serve as an effective complement to existing approaches.

\end{abstract}

\section{Introduction}

Reinforcement Learning (RL) is an area of machine learning that seeks to create agents that learn what to do in uncertain, stochastic environments. That is, RL agents learn how to map situations to actions with the goal of maximizing some type of reward \cite{sutton_reinforcement_2018}. A core tenant of RL that is touted as a key benefit is that agents learn to do this autonomously, without human intervention, in an unsupervised fashion. The typical engineering workflow in the design of RL agents is 1) a reward function is designed, 2) input and output channels are selected, and 3) an RL learning algorithm is designed \cite{abel_agent-agnostic_2017}. And once the RL pipeline has been designed and implemented, little if any human intervention is needed during the learning process.

However, for real-world situations, where the high complexity of the world cannot easily be modeled in simulation environments, autonomous learning is often not feasible, primarily because of three key issues: the difficulty in specifying reward functions \cite{ng_algorithms_2000}, the constraints on exploration due to the presence of potentially catastrophic outcomes for agents and humans alike \cite{garcia_comprehensive_2015}, and difficulty in communicating goals to RL agent so as to avoid being misinterpreted \cite{amodei_concrete_2016}. As a result, many researchers have proposed introducing a human in the loop of training RL agents, to varying degrees of success \cite{abel_agent-agnostic_2017,knox_augmenting_2011,knox_reinforcement_2011,liu_reinforcement_2002,mathewson_simultaneous_2016,abbeel_apprenticeship_2004}.

These examples not only have elements of machine learning, but also elements of human-computer interaction (HCI). First coined in 1983 by Card et. al \cite{card_psychology_1983,card_keystroke-level_1980}, HCI as a discipline is concerned with the design, evaluation and implementation of interactive computing systems for human use and with the study of major phenomena surrounding them \cite{hewett_acm_1992}. Similarly, there has been some work related to having RL agents learn user preferences \cite{veeriah_face_2016,christiano_deep_2017}. For example, Veeriah et. al \cite{veeriah_face_2016} demonstrated that an RL agent can learn a human user's intent and preferences via ongoing interactions where the human user provides feedback through facial expressions, as well as, through explicit negative rewards. Collectively, the interface between machine learning and human computer interaction is referred to as Interactive Machine Learning (iML). In this paper, I explore combining Reinforcement Learning techniques with spatial interfaces to create agents that learn a value function that relates a user's body language, specifically a thumbs up or thumbs down gesture, to expectations of future rewards.

\subsection{Interactive Machine Learning}
Interactive Machine Learning (iML) is a subfield of Machine Learning that creates systems that operate in environments alongside human operators, where the human and the machine collaborate to achieve a task and both the human and ML system can be agents in the environment \cite{holzinger_interactive_2016}. Many times, iML is used to extend or amplify the capabilities of a human in cognitive or physical ways. For this to be successful, the machine must adapt to the users' intentions and preferences, learning about their human counterparts' behavior. In much of the current research in iML, a user must convey feedback in some way to the iML agent.

One popular method is directly, such as through the click of a mouse, or through spoken word. This is known as explicit human feedback. Implicit human feedback on the other hand, is a mechanism through which a human can guide an iML agent's learning process through subtle cues, such as body language. And yet a third type of feedback is known as mixed human feedback which combines explicit and implicit feedback \cite{boukhelifa_evaluation_2018}. Although all three forms of feedback impose some cognitive load on the human, explicit feedback carries the heaviest cognitive burden, especially in real-world settings where humans have additional cognitive loads due to the environment. The main objective of this work is to extend the current state of the art in iML by demonstrating that a learning agent can learn a human user's behavior and adapt to the human's preferences with a reduced amount of explicit human feedback in a mixed feedback setting.

\subsection{Related Work}
With the growth of the machine learning field over the last decade, there has been much effort in the community to create successful interactions between humans and machine learning systems with the goal of increase performance across a myriad of tasks. Many approaches to human-in-the-loop machine learning have focused primarily on agents learning from humans via explicit rewards. For example, Thomaz and Breazel \cite{thomaz_teachable_2008} used a simulated RL robot and had a human teach the robot in real-time to perform a new task. They presented three interesting findings: 1) humans use the reward channel for feedback, as well as for future-directed guidance; 2) humans exhibited a positive bias to their feedback using the signal as a motivational channel; and 3) humans change their behavior as they develop a mental model of the robotic learner.

More recently, Knox \& Stone \cite{knox_teaching_2013,knox_framing_2015} introduced the TAMER framework (Training an Agent Manually via Evaluative Reinforcement). TAMER's system for learning from human reward is novel in three key ways: 
\begin{itemize}
    \item TAMER addresses delays in human evaluation through credit assignment
    \item TAMER learns a predictive model of human reward
    \item and at each time step, TAMER chooses the action that is predicted to directly elicit the most reward, disregarding consideration of the action's effect on future state (i.e., in reinforcement learning terms, TAMER myopically values state-action pairs using a discount factor of 0).
\end{itemize}
According to Knox \& Stone, ``TAMER is built from the intuition that human trainers can give feedback that constitutes a complete judgement on the long-term desirability of recent behavior.'' One drawback of this method, however, is that when the user needs to modify the agent's behavior, the model would need to be changed, for example, via additional rewards from the user.

Another interesting approach is by Christiano et. al \cite{christiano_deep_2017}. Here, the agent is trained from a neural network known as the ‘reward predictor’, instead of the classical RL approach of using the rewards it collects as it explores an environment. There are three processes running in parallel:
\begin{itemize}
    \item A RL agent explores and interacts with its environment
    \item Periodically, a pair of 1-2 second clips of its behavior is sent to a human who is then asked to select the best one indicating steps toward fulfilling the desired goal
    \item The human’s choice is used to train a reward predictor, which is then used to train the agent
\end{itemize}
 They showed that overtime, the agent learns to maximize the reward from the predictor and improve its behavior according to the human’s preferences. However, this methodology introduces substantial lag between the human feedback and the agent's learning.
 
 Another approach by Veeriah et. al \cite{veeriah_face_2016} uses body language as one of the drivers of learning. In contrast to the other approaches mentioned, this approach is concerned with designing a general, scalable agent that would allow a human user to change the agent's behavior according to preferences with minimal human feedback. In this work, they used facial expressions to provide the feedback, not as a channel for control, but rather as a means of valuing the agent's actions. 
 
 This work is an extension of the work by Veeriah et. al \cite{veeriah_face_2016} to hand gestures captured by a 3D spatial interface.

\subsection{Problem Statement}
There are many domains for which human and computing systems, enabled by some form of machine learning, work collaboratively to achieve various tasks. Once such domain of human computer interaction is in the realm of prostheses, whereby a human operator controls an electronic prosthetic limb \cite{li_pca_2018,edwards_application_2016}. Grip selection, for example, is one of the main tasks a human operating a prosthetic limb performs. Performing grasping of common objects with the prosthesis requires choosing the correct grip from an array of various configurations. Most systems cycle through grips via some feedback mechanism, typically a roll or pronation/suprenation \cite{resnik_advanced_2012}.

Thus, a grip selection task environment was created in order to evaluate the machine learning system. This problem is taken from the real-world task of selecting an appropriate grip pattern for grasping a given object by a user that is operating a prosthetic arm, as mentioned above. Typically for most prosthetic arms, there are a set of $n$ discrete grips and depending on the type of object or situation, the correct grip is defined according to the user's preference. These preferences are normally setup by a clinician and must be periodically revisited as the user gains experience with the prosthesis \cite{resnik_advanced_2012}. For this experiment, the agent must select the correct grip for a random object presented, move forward, and grasp the object. The episode ends when the agent successfully grasps the object. In this way, the agent learns the user's gripping preferences in an ongoing and online fashion, reducing the need for multiple clinical visits.

\section{Reinforcement Learning Algorithm Background}

Reinforcement Learning is a branch of Machine Learning that allows agents to decide what to do, that is, how to map situations to actions with the goal of maximizing a numerical reward signal. RL agents typically interact with an environment, either real-world or simulated, gaining experience with each interaction, and improving their performance, as measured by the reward signal. Mathematically, Markov Decision Processes (MDPs) are used to describe RL problems. There are several algorithms that solve MDPs, many in optimal ways. The RL learning algorithm that was investigated for the grip selection task was SARSA($\lambda$), which is an on-policy temporal difference (TD) control algorithm for MDPs \cite{sutton_reinforcement_2018}.

\subsection{Markov Decision Processes}
A Markov Decision Process is a tuple $(S, A, {P_sa}, \gamma, R)$, where:
\begin{itemize}
    \item $S$ is a set of \textit{states}, $S \in \{s_0, s_1, \dots s_m\}$
    \item $A$ is a set of \textit{actions}, $A \in \{a_0, a_1, \dots a_m\}$
    \item $P_{sa}$ are the state \textit{transition probabilities}
    \item $\gamma \in [0, 1)$ is the \textit{discount factor}, and controls the influence of future rewards in the present
    \item $R:S \times A \mapsto \mathbb{R}$ is the \textit{reward function}, which returns a real value every time the agent moves from one state to the other due to an action
\end{itemize}
For each state $s \in S$ and action $a \in A$, $P_{sa}$ is a probability distribution over the state space and gives the distribution over what states the system will transition to if action $a$ is taken when the system is in state $s$. Starting from an initial state, $s$, the agent can choose to take an action $a \in A$. The state of the MDP then randomly transitions to a successor state $s'$. The successor state is drawn according to $s' \sim P_{sa}$. In MDPs, the transition model depends on the current state, the next state, and the action of the agent. This process happens sequentially over time, with each new action, generating a new successive state, which leads to further actions.

The reward function, $R(s,a)$, is the primer driver of RL algorithms. Having the correct $R(s,a)$ can make or break RL algorithms in that the reward function is essentially the ``teacher'' in the learning algorithm. It indicates to the RL algorithm which state-action pairs are more desirable than others by assigning values. For example, one methodology is to assign positive value to state-action pairs that are desirable and a zero or negative values to those that don't matter or are not as desirable.

Thus, the goal in reinforcement learning is to maximize the reward over time, choosing states via actions that increase the return over time, and avoiding states that decrease the return over time. If we start at time t in state $s_t$ and choose an action $a_t$, where $s_i$ and $a_i$ are states and actions in a sequence, and $t = 0, 1, 2, \dots$, we can represent the the MDP as follows:

\begin{equation}
    s_t \xrightarrow[]{a_t} s_{t+1} \xrightarrow[]{a_{t+1}} s_{t+2} \xrightarrow[]{a_{t+2}} s_{t+3} \dots
\end{equation}

Thus, the total reward at time $t$ if we take $a_0, a_1, \dots$ actions and visit $s_0, s_1, \dots$ states over time $t+1, t+2, t+3, \dots$ is given by:

\begin{align}
  R_t(s,a) &= R_{t+1}(s_t, a_t) + \gamma R_{t+2}(s_{t+1},a_{t+1}) + \gamma^2 R_{t+3}(s_{t+2},a_{t+2}) + \cdots \nonumber \\
  R_t(s,a) &= \sum_{i=1}^\infty \gamma^{i-1}R_{t+i}(s_{i-1},a_{i-1})
\end{align}

Thus, the goal would be to maximize the expected value of the total reward:

\begin{equation}
E[R_t(s,a)]
\end{equation}

The solution to this problem is to find a \textit{policy}, $\pi$, which returns the action that will yield the highest reward for each state. A policy is any function that maps states to actions: $\pi : S \mapsto A$. When we \textit{execute} a policy $\pi$ when in a state $s$, the agent takes action $a'=\pi(s,a)$. There can potentially be many policies to choose from, but only one can be considered an optimal policy, which is denoted by $\pi^*$ and yields the highest expected reward over time, which is call the \textit{action value function}. Thus, an action value function for a policy $\pi$ is:

\begin{equation}
q^\pi(s,a)=E[R_t|S_t=s, A_t=a, \pi]
\end{equation}

\noindent where $q^\pi(s,a)$ is the expected sum of discounted rewards when starting in state $s$ and taking action $a$ according to policy $\pi$.

The goal is to find an \textit{optimal policy}, $\pi^*$ that maximizes the action value function:

\begin{equation}\label{greedyPolicy}
\pi_{(s,a)}^* = \underset{\pi}{\text{argmax}}\ q^\pi(s,a)  
\end{equation}

For most practical applications of RL, $q^\pi(s,a)$ is not known and must be learned by interacting with the environment. This type of RL agent, one that learns the action-value function not from a transition model for the environment, but rather from direct interaction with the environment, is called a model-free RL agent.

\subsection{\texorpdfstring{SARSA($\lambda$)}{SARSA(λ)}}
In RL, rewards are viewed as short-term signals of the quality of an action, where as the action value function, $Q^\pi(s,a)$ represents the long-term value of a state-action pair. Temporal difference (TD) learning is a class of model-free methods that estimates $Q^\pi$ as the agent interacts with the environment. The agent samples transitions and then updates the estimate of $Q^\pi$ using observed and the estimate of the values of the next action. Typically, the agent makes these observations and updates the action value function at every time step, according to the following update rule:

\begin{equation}\label{q_update}
    Q(S_t, A_t) \xleftarrow{} Q(S_t, A_t) + \alpha \delta_t
\end{equation}

\noindent where $Q$ is an estimate of $q^\pi$, $\alpha$ is the step size, and $\delta_t$ is the TD error. SARSA is a TD learning algorithm that samples states and actions using an $\epsilon$-greedy policy and then updates the $Q$ values using equation \ref{q_update} and a $\delta_{t}$ as follows:

\begin{equation}\label{d_delta}
    \delta_t = R_{t+1} + \gamma Q(S_{t+1},A_{t+1}) - Q(S_t,A_t)
\end{equation}

\noindent The term $R_{t+1} + \gamma Q(S_{t+1},A_{t+1})$ is called the target. It consists of the reward plus the discounted value of the next state and next action.

SARSA is known as an on-policy method which arises from the fact that the behavior policy $u$ is the same as the target policy $\pi$. That is, the TD target in SARSA consists of $Q(S_{t+1},A_{t+1})$, where $A_{t+1}$ is sampled using $\mu$. The target policy $\pi$, is used to compute the TD target. Although On-policy methods may result sub-optimal policies in certain instances, it has been shown that policies learned in on-policy methods tend to be safer when the risks are greater because SARSA takes the action selection into account \cite{sutton_reinforcement_2018}.

Eligibility traces, a method of including information about not just the current time step, but information from multiple time steps, are a key mechanisms in reinforcement learning. For example, in TD($\lambda$) algorithm, the $\lambda$ refers to the use of an eligibility trace. Almost any TD method, such as SARSA, can be combined with eligibility traces to obtain a more general method that may learn more efficiently \cite{sutton_reinforcement_2018}. In this study, SARSA($\lambda$) is used to improve efficiency of the learning agent. 

\subsection{Function Approximation using Tile Coding}
In order to implement equations \ref{q_update} and \ref{d_delta}, an estimate of $Q$ must be maintained and updated. Since there is no analytical way of expressing $Q$ as a function, a method of funciton approximation is often used. Formally, function approximation is a technique for ``representing the value function concisely at infinitely many points and generalizing value estimates to unseen regions of the state- action space'' \cite{sherstov_function_2005}. \textit{Tile coding} is a linear function approximation method that is flexible and computationally efficient. In tile coding, the variable space, which is typically composed of states and actions, is partitioned into tiles, with each partition called a tiling. The method uses several overlapping tilings and for each tiling, maintains the weights of its tiles. The approximate value of an input is found by summing the weights of the tiles, one per tiling, in which it is contained. Given a training example, the method adjusts the weights of the involved tiles by the same amount to reduce the error on the example \cite{sherstov_function_2005}.

\subsection{Action Selection Policy}
As the action-value function is being learned via a function approximation method such as Tile Coding, the agent needs a method to select an action from the current set of actions whose action-value is known. Equation \ref{greedyPolicy} always chooses the action that yields the largest action value. This is known as \textit{greedy} action selection. But since the agent learns this function through experience, there may be other actions that the agent may never see that will yield better values. In order to remedy this, three popular action selection strategies are often used: $\epsilon$-greedy, $\epsilon$-soft, and soft-max \cite{sutton_reinforcement_2018}. 

\subsubsection{\texorpdfstring{$\epsilon$-greedy}{ε-greedy}}
In $\epsilon$-greedy, the action with the highest value is chosen most of the time. However, every so often, an action is selected at random with $\epsilon$ probability. The action is selected uniformly, independent of the action values. This guarantees that every action will be explored sufficiently to find the optimal policy. Larger values of $\epsilon$ favor more exploration of actions, while smaller values favor exploitation of the greediest action. This is known as the exploration vs exploitation trade-off.

\subsubsection{\texorpdfstring{$\epsilon$-soft}{ε-soft}}
$\epsilon$-soft action selection is similar to $\epsilon$-greedy, except that the best action is selected with probability $1 - \epsilon$ and a random action is selected the rest of the time.

\subsubsection{softmax}
A limitation of $\epsilon$-greedy and $\epsilon$-soft is when selecting random actions, they are selected uniformly. The worst action to take is selected with equal probability to the best action to take. Enter softmax. In softmax action selection, first, a rank or weight is assigned to each of the actions, according to their action-value estimate. Then, a random action is selected with regards to the weight associated with each action. This means that as the action-value function is learned, poor actions would be less likely to be chosen. This is useful, particularly the worst actions have unfavorable consequences.

For this study, an $\epsilon$-greedy action selection strategy was used for its simplicity.

\section{System Design and Experimental Setup}
While having a working prosthetic to develop a solution for learning grip preferences using reinforcement is desirable, prosthetic devices are both expensive and challenging to configure and interface with. Thus, in order to carry out the agent design and experimentation in a low-cost way that would be accessible by other researchers or iML practioners, a simulation environment was created. The simulation consisted of a grip selection task with object grasping and was created using two popular and well supported open source projects, Pybullet and OpenAI gym.

\subsection{Simulation Environment}
\begin{figure}[tb!]
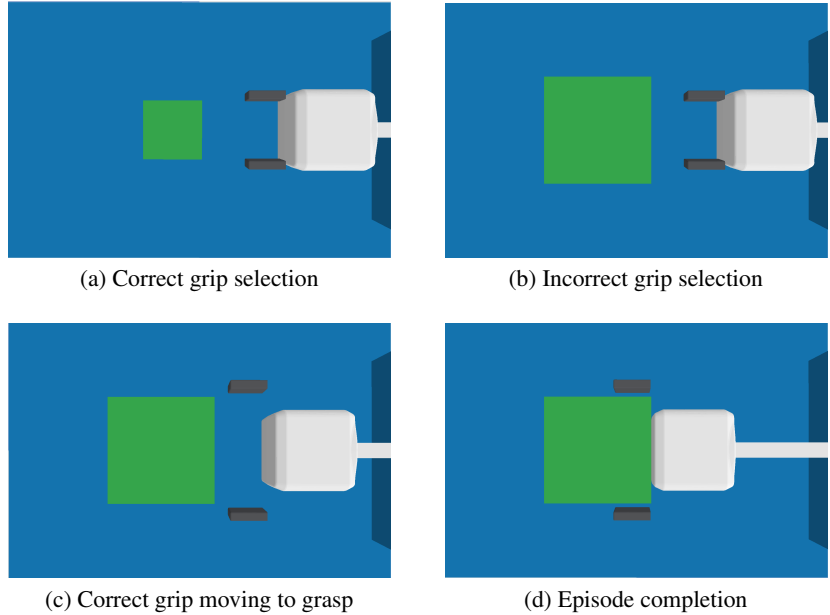

\centering
\subfloat[Correct grip selection]{\includegraphics[width = 2in]{gripper1.png}}\hspace{0.25in}
\subfloat[Incorrect grip selection]{\includegraphics[width = 2in]{gripper2.png}}\\
\subfloat[Correct grip moving to grasp]{\includegraphics[width = 2in]{gripper3.png}}\hspace{0.25in}
\subfloat[Episode completion]{\includegraphics[width = 2in]{gripper4.png}} 
\caption{Simulation Environment in Pybullet using the OpenAI gym interface. In (a), the arm is in the starting position called the ``grip changing station''. It has selected the correct grip for the object size. In (b), the incorrect grip has been selected, although the arm still remains in the grip chaning station. In (c), the correct grip for the object size as been enabled and the arm is moving forward to grasp the object. Finally, (d) illustrates the completion of an episode.}
\label{sim_env}
\end{figure}

The grip selection task was simulated using the Bullet physics simulator\footnote{\url{https://pybullet.org/wordpress/}} and the OpenAI gym interface\footnote{\url{https://gym.openai.com/}}. A robotic arm with three degrees of freedom was modeled and imported into the Bullet environment. The robotic arm consists of two grippers, each with one degree of freedom, and the arm portion with one degree of freedom. The arm is attached to a table on which objects of variable size are placed, one at a time. Figure \ref{sim_env} shows different scenarios that are possible in the simulation.

Although the Bullet environment offers an API that can be used to control the environment to do things such as restart an episode, control the arm, or add an object of random size, the OpenAI gym interface provides a consistent, common API that can be used to develop agents that can operate in multiple environments, as opposed to developing environment specific agents. Thus, the OpenAI gym interface was used to abstract the specific environment details of the grasping environment away from the agent. This allows for de-coupling of the agent development from the environment development and allows other researchers to use the grasping environment with their own custom agents. By using the Bullet physics engine along with the OpenAI gym environment, several experiments were carried out with little effort.

\subsection{Spatial Interface Valuing}
In order to have an agent learn from a human, some form of feedback is required. In this work, a learning agent perceives a value of its own behavior from human hand gestures given via a spatial interface, which I term Spatial Interface Valuing (SIV). In order to capture hand gestures, a Leap Motion Controller\footnote{\url{https://www.leapmotion.com/}}(LMC) is used as the spatial interface device. The LMC provides tracking information for the left and right hands, as well as simple gesture recognition. The gestures of interest for this study are a thumbs up or thumbs down. Unfortunately, that is not one of the stock gestures implemented. Instead, the roll ($\rho$) of the right hand was used to determine whether the user was indicating a thumbs up or thumbs down. A simple piece-wise function was used to determine the state of the hand: thumbs up (1.0), thumbs down (-1.0).

\begin{equation}
    hand\_state(\rho) = 
    \begin{cases}
    1.0 & \text{if $-45 < \rho < -135$} \\
    -1.0 & \text{otherwise}
    \end{cases}
\end{equation}

\begin{figure}[tb!]
    \centering
    \includegraphics[scale=0.35]{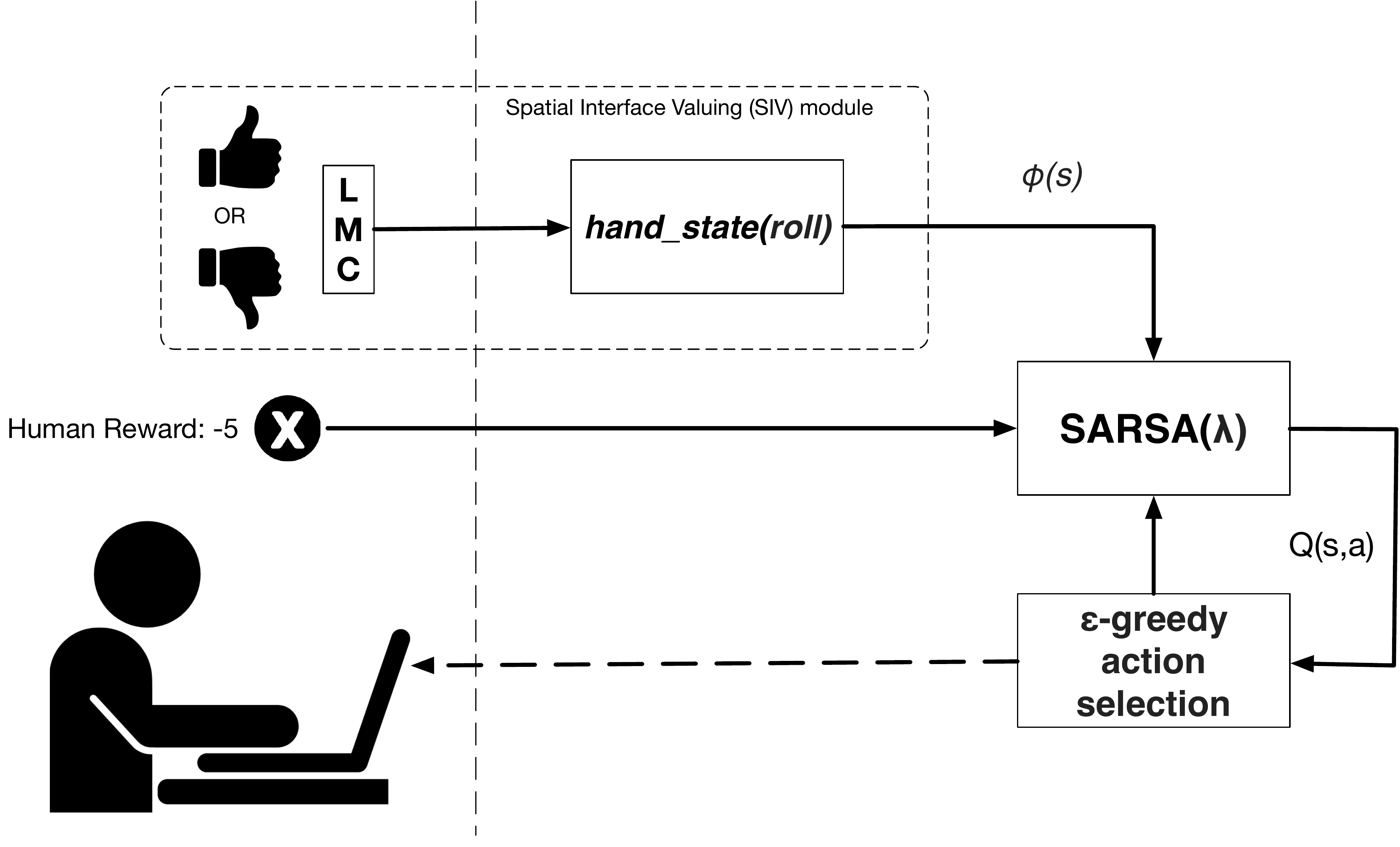}
    \caption{Overview of the SIV agent setup}
    \label{siv_setup}
\end{figure}

\subsection{Experimental Setup}\label{exp_setup}
The experimental setup consists of a user observing the simulated grip-selection task and assisting the agent during the training process by 1) signaling a thumbs up or thumbs down via the spatial interface to signal approval or disapproval of the agents behavior and 2) pushing the space bar on the keyboard when the agent was not behaving as expected, giving the agent a negative reward. A SARSA($\lambda$) agent both with and without SIV, as well as baseline agent was evaluated, each for 3 runs with 15 episodes per run. All of the experiments were carried out without the user knowing which of the three agents was currently being evaluated, i.e. in a blind setting. Additionally, the order in which the experiments were carried out was randomly selected over the combination of runs, episodes, grips, object sizes, and agent. Figure \ref{siv_setup} shows an overview of the experimental setup, with the only difference between agent the SIV/No SIV agents is the feedback from the hand state. 

\paragraph{Note:} Three distinct simulation environments were created for this study, a baseline environment for comparison, one environment with SIV feedback, and one environment without SIV. The experimental setup (excluding the SIV module) is the same, with the only difference being the inclusion (or exclusion) of the SIV module. For both simulation environments, the agent makes observations of the state space every \textit{one-tenth of a second} and must take an action at every time step.

\begin{table}[tb!]
\centering
\caption{State Spaces for each agent under test: SIV vs no SIV}\label{tab1}
\begin{tabular}{| C | D |}
\hline
Agent Type &  State Space Variables\\
\hline
baseline & current grip size, current object size, bias \\
with SIV & current grip size, hand state, bias \\
without SIV & current grip size, bias \\
\hline
\end{tabular}
\label{tab:1}
\end{table}

\subsubsection{State Space}
As mentioned above, three SARSA($\lambda$) \cite{sutton_reinforcement_2018,rummery_-line_1994} agents, a baseline, one with SIV and one without SIV, were implemented to determine how well the agents learned a user's preferences for the grip selection task. Table \ref{tab1} describes the state space design for all three agents. The SARSA($\lambda$) baseline agent uses the size of the object and the current grip, along with a bias term as the state space vector $\phi (s)$:

\begin{equation}
    \phi(s) = [current\_grip\_size, current\_object\_size, bias]
\end{equation}

This state space was chosen in order to allow the agent to learn the best grip for each object, since both values are known form within the simulation, they can be observed and are used to establish a baseline level of performance with which to compare the SIV an No SIV enabled agents. The maximum grip size, maximum object size, number of grips, and number of objects are all parameters that can be configured prior to running each episode. 

For the agent enabled with SIV, which is ideally for deployment in a real prosthetic device, the grip size is known to the agent, but for real-world grasping tasks, object size is not known. This is where SIV steps in. Having knowledge of the user's gesture takes the place of exact knowledge of the object size (as in the baseline agent). Thus, the state vector for the SIV agent is:

\begin{equation}
    \phi_{SIV}(s) = [current\_grip\_size, hand\_state, bias]
\end{equation}

And lastly, the state space for the agent that does not use SIV as a feedback channel, only has knowledge of the current grip.

\begin{equation}
    \phi_{No SIV}(s) = [current\_grip\_size, bias]
\end{equation}

\subsubsection{Action Space} Modeled after the grip selection task in \cite{veeriah_face_2016}, the complete action space for the agents consists of the following:

\begin{equation}
    \mathcal{A}_{complete}(s) = \{grip_1, grip_2, \dots grip_n, \leftarrow, \rightarrow\}
\end{equation}
where the first $1 \dots n$ actions are grip selections amongst the $n$ grips. The remaining actions $\{ \leftarrow, \rightarrow \}$, when taken, move the arm one step closer to the object ($\leftarrow$) or one step closer to the grip changing station ($\rightarrow$). However, the actions that are available to the agent depending on the position of the arm relative to the grip changing station, the object, and the reward. When the arm is in the grip changing station, the actions available are:
\begin{equation}
    \mathcal{A}_{grip\_change}(s) = \{grip_1, grip_2, \dots grip_n, \leftarrow\}
\end{equation}

Once the grip has been selected and the agent moves forward one step towards the object, the actions available are:

\begin{equation}
    \mathcal{A}_{move}(s) = \{\leftarrow, \rightarrow\}
\end{equation}
And lastly, if the agent receives an explicit negative reward from the human user, the only available action is to return to the grip changing station:
\begin{equation}
    \mathcal{A}_{return}(s) = \{ \rightarrow\}
\end{equation}

\begin{table}[t!]
\centering
\caption{Hyperparameters for SARSA($\lambda$) agents}
\begin{tabular}{ C | C }
\textbf{Parameter} &  \textbf{Value}\\
\hline
$\lambda$ & 0.0 \\
step size & 0.5 \\
$\gamma$ & 1.0 \\
$\epsilon$ & 0.1\\
\end{tabular}
\label{tab:2}
\end{table}

\section{Experiments}

\subsection{Two objects with Four Grips}
In order to validate SIV as an RL method that can effectively learn a user's preference during grip selection, several trials of a grip selection task were carried out with a user in a blind setting, as mentioned in Section \ref{exp_setup}. At the beginning of each episode, the environments were setup to randomly select one of two object sizes: a small object, much smaller than the largest grip size, and a large object, approximately the size of the largest grip. The number of grips for the grip selection task was set to four for all episodes, requiring that the agent learn the users grip preference in two distinct grasping situations. The experiments were formulated as an episodic MDP with 0 discount ($\gamma$ = 1.0), 0 reward at every time step and a 0 reward for completing the episode. The hyperameters for SARSA($\lambda$) agents are show in table \ref{tab:2}.

\begin{figure}
\centering
\begin{tabular}{cc}
  \includegraphics[width=60mm]{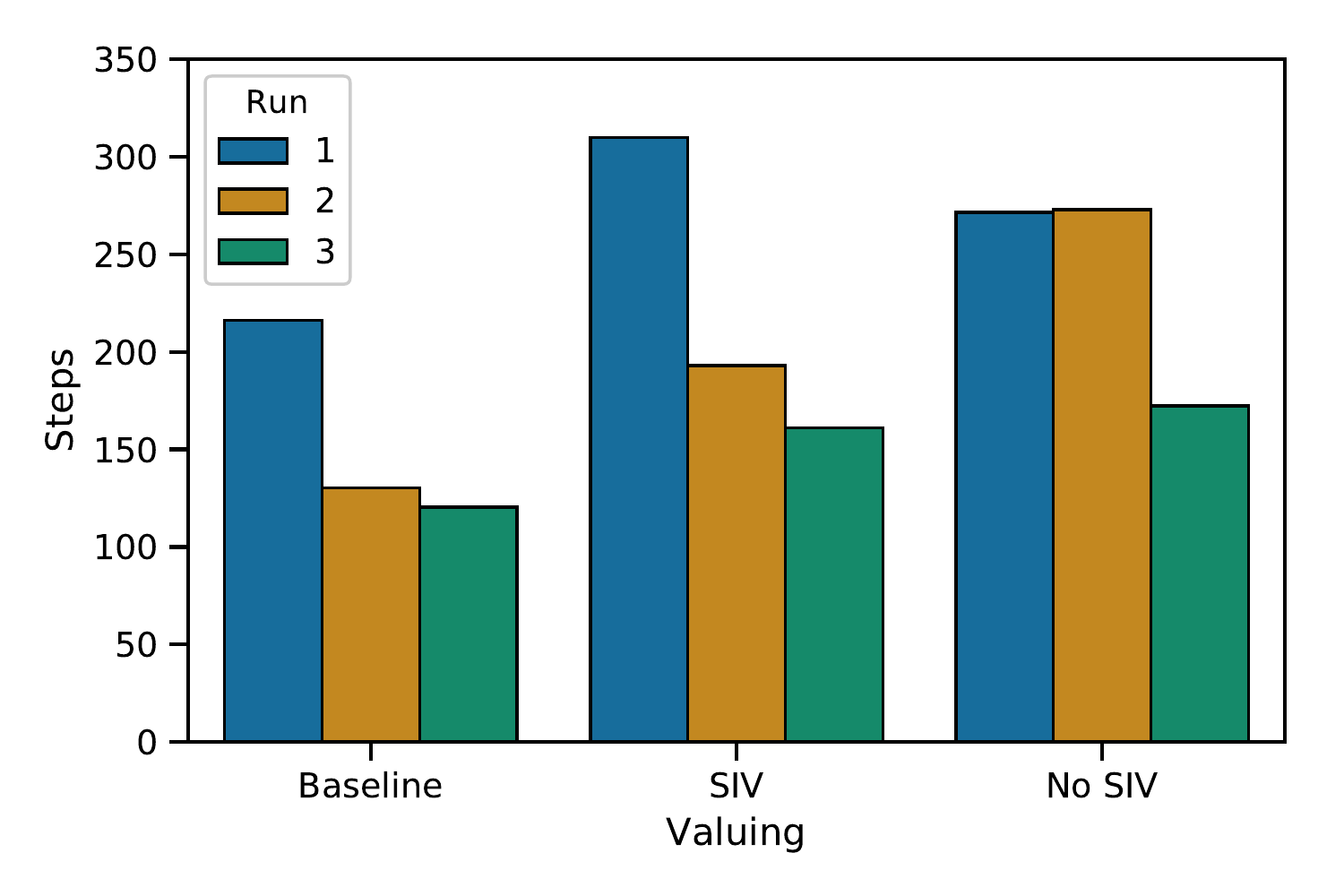} &   \includegraphics[width=60mm]{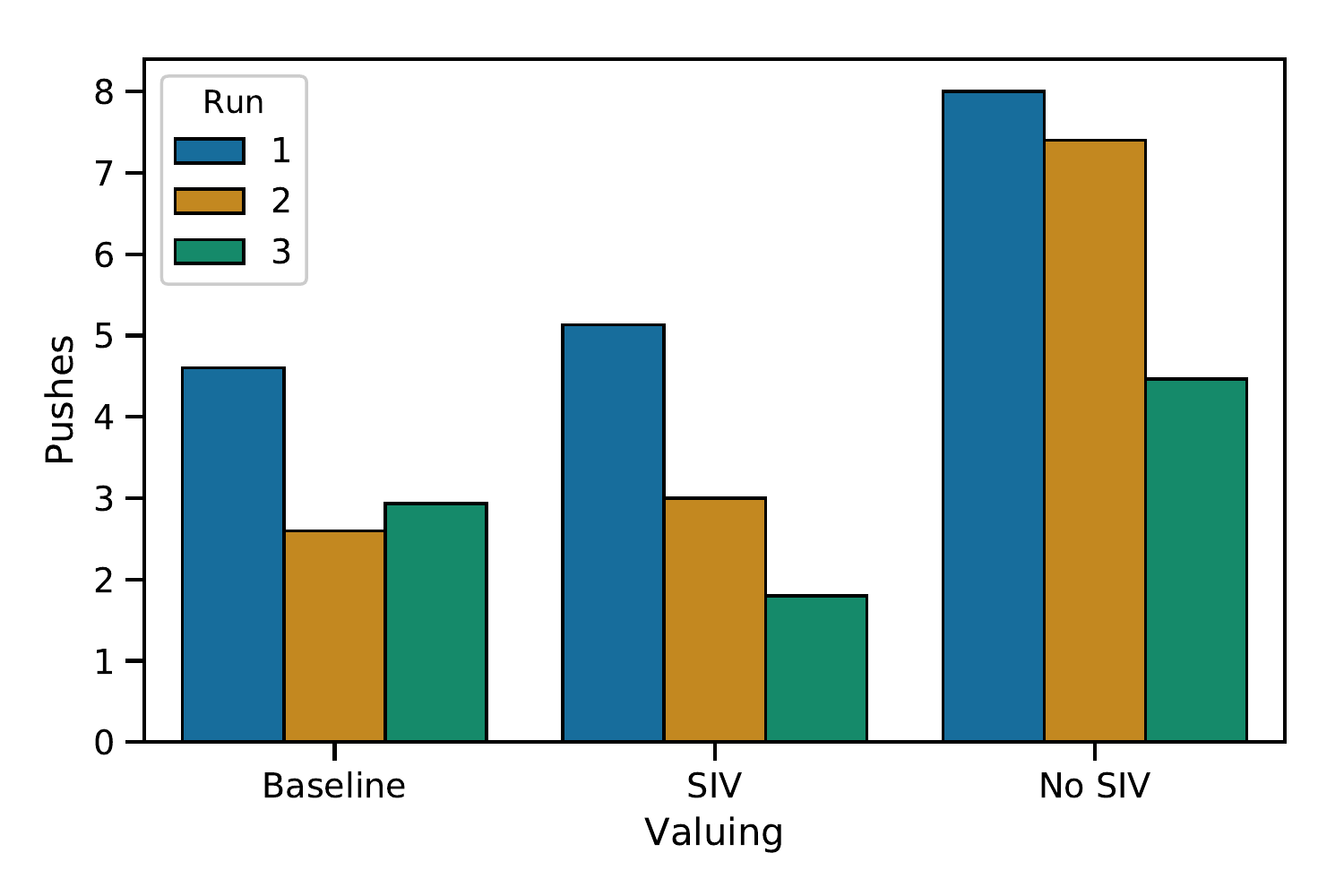} \\
(a) Average steps per run & (b) Average pushes per run \\[6pt]
 \includegraphics[width=60mm]{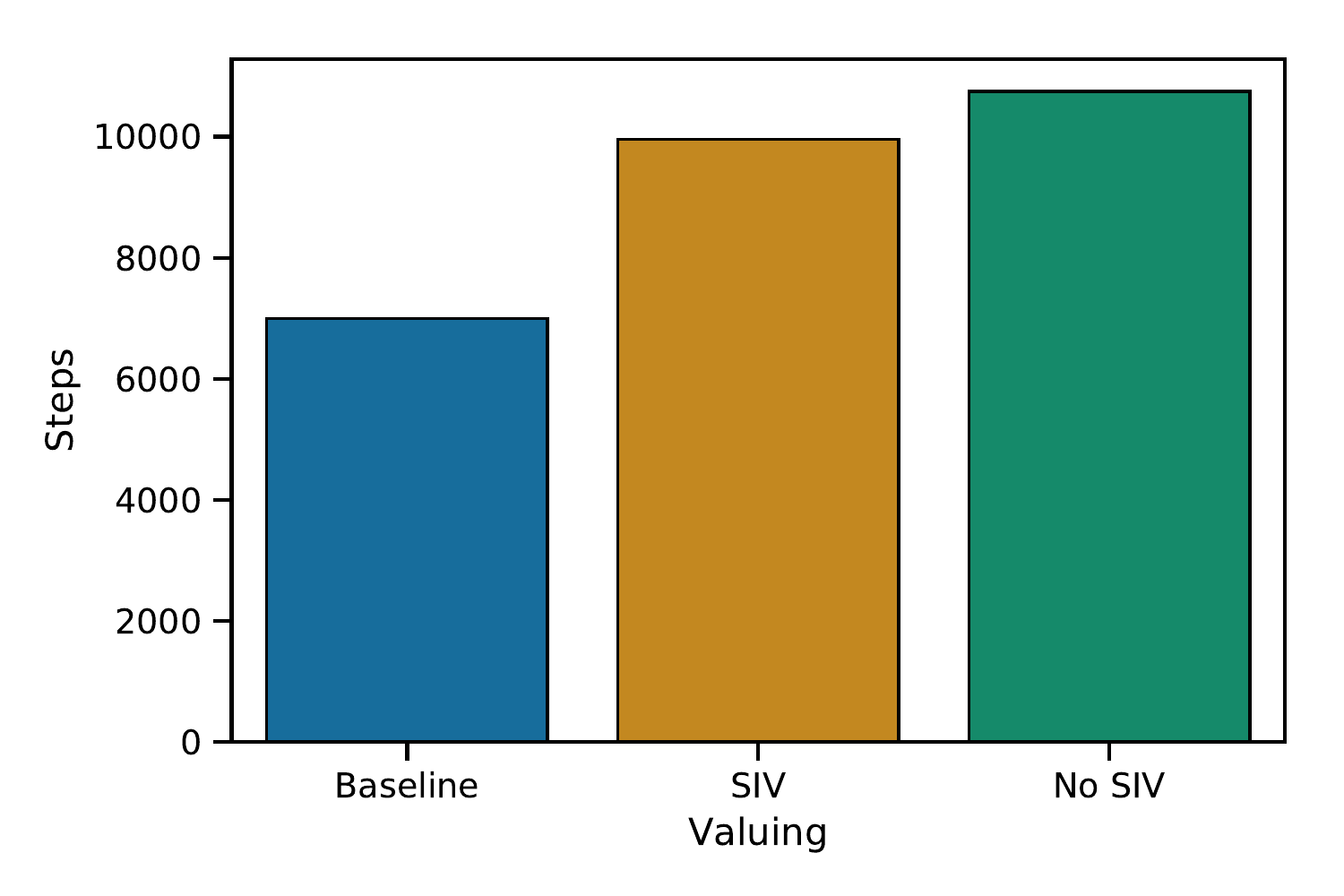} &   \includegraphics[width=60mm]{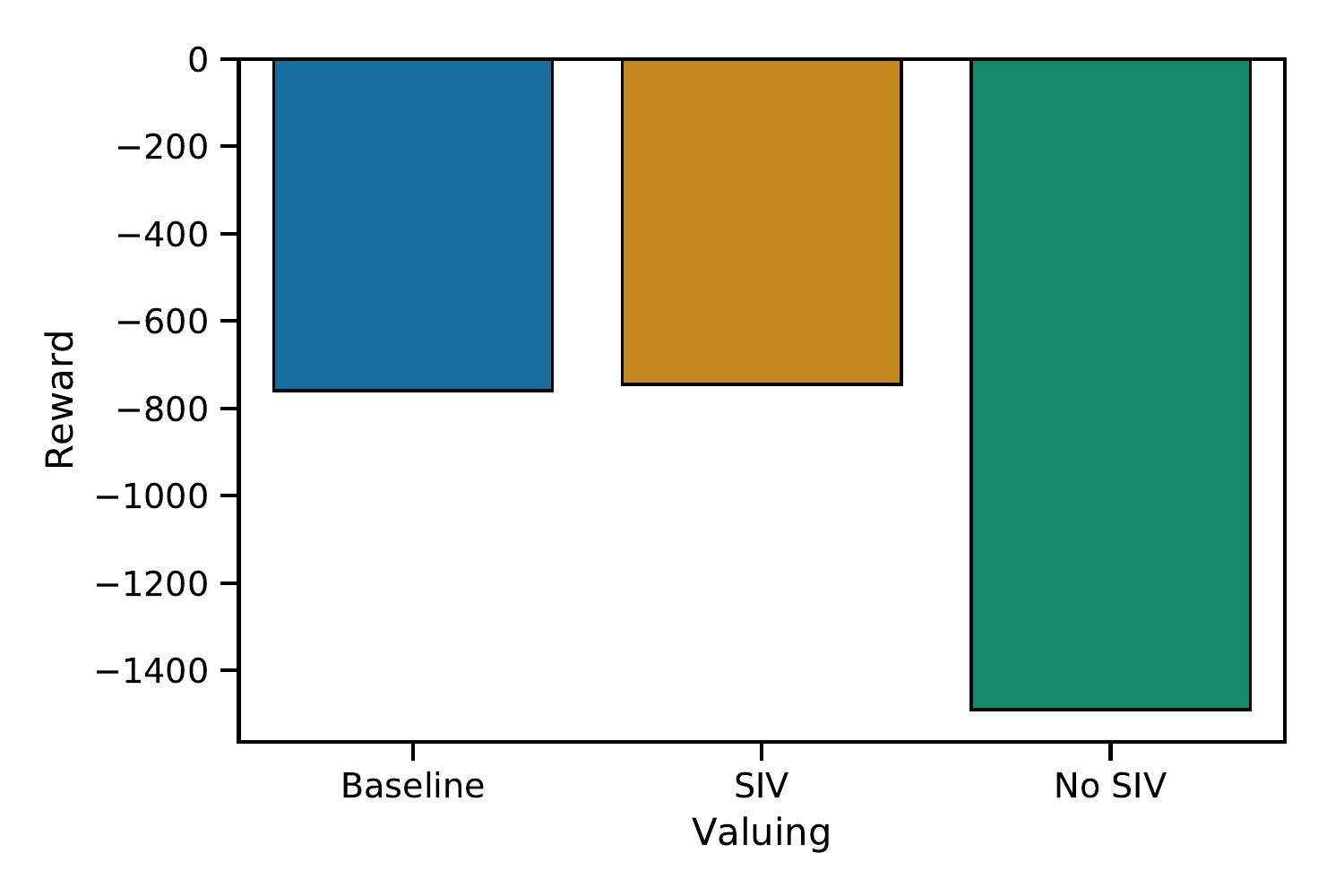} \\
(c) Total steps per method & (d) Total reward per method \\[6pt]
\multicolumn{2}{c}{\includegraphics[width=75mm]{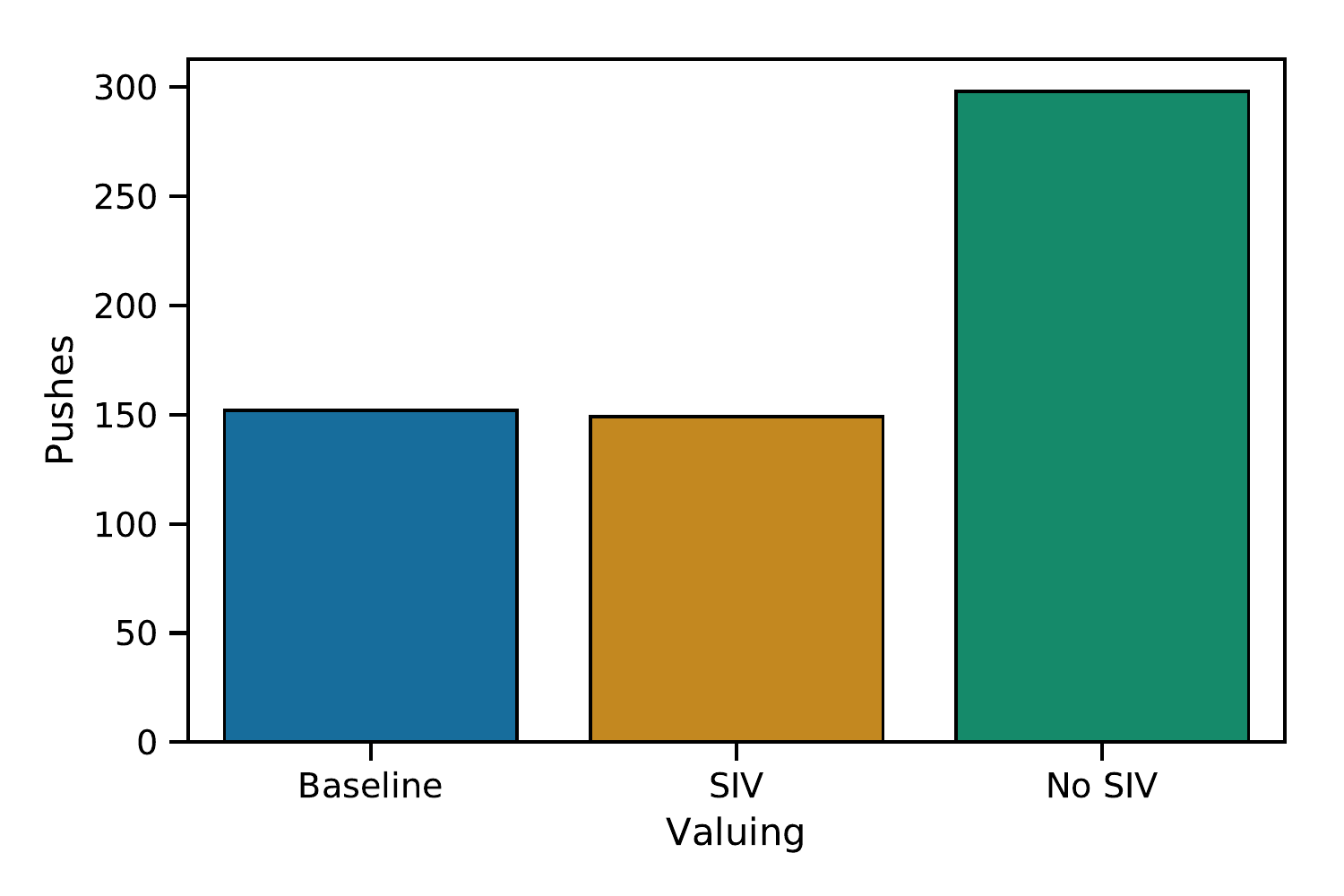} }\\
\multicolumn{2}{c}{(e) Total pushes per method}
\end{tabular}
\caption{Results from evaluating agents using SIV against a Baseline agent with full knowledge of the current grip and object size, and an agent not using SIV with knowledge of only the current grip.}
\label{results}
\end{figure}

\subsection{Experimental Results}
The results for the experiment are shown in Figure \ref{results}. Overall, the baseline agent performs the best when considering the average steps per episode (Figure \ref{results}(a)). This is to be expected, because once the baseline agent learns the user preference, it can execute those preferences directly because the baseline agent has knowledge of both the object size and current grip and has learned the action value function and thus, the optimal policy. Once trained, it does not need to rely on the human for any form of input. However, observing the performance of the agent using SIV, although on average it took more steps to complete an episode than the baseline, it out performed the agent without SIV on two of the three runs. This may be due to the human user themselves getting accustomed to providing the feedback to all of the agents.

Another interesting finding is that both the baseline agent and the agent with SIV learned with approximately the same number of average reward button pushes, where as the agent learning the grip selection task without SIV needed many more button pushes to successfully complete the runs. This indicates that the agent with SIV learns as quickly as the baseline agent, while the agent without SIV takes longer to learn the user's preference. This is likely because the agent without SIV only has the reward channel to try to learn the human users preference.  This is shown in Figure \ref{results}(b) and Figure \ref{results}(e). Similarly, the amount of total reward needed and total number of button pushes needed (Figure \ref{results}(d) and \ref{results}(e)) for the baseline and SIV agents is approximately half what is needed for the agent without SIV.

\section{Conclusions and Discussion}
This study introduces a new approach called \textit{Spatial Interface Valuing} that uses a 3D spatial interface device to adapt an agent to a user's preferences. I showed that SIV can produce substantial performance improvements over an agent that does not use SIV. Through SIV, the agent learns to adapt its own performance and learn user preferences through gestures, reducing the amount of explicit feedback required. SIV delivers implicit feedback from a user's hand gestures to an agent, allowing the agent to learn much more quickly and require less explicit human generated reward. The SIV agent learned to map hand gestures to user satisfaction, codifying satisfaction as a action value function using temporal methods for Reinforcement Learning. This technique is task agnostic and I believe it will easily extend to other settings, tasks, and forms of body language.

\section{Future Work}
The current results are encouraging, however, more work can be done on several fronts. This study was limited to studying the SARSA($\lambda$) RL algorithm, an on-policy agent that uses tile coding, a function approximation technique, to estimate the action value function and learn a policy that encodes user preferences. It may be useful to study other types of agents such as off-policy agents using Q-learning, for example. Additionally, deep learning techniques such as Deep Q Networks \cite{mnih_human-level_2015-1} and DeepSARSA \cite{zhao_deep_2016} may perform better. Additionally, simply extending the size of the experiment by conducting more runs with more users, different hyper-parameters and random seeds, and different object size/grip number combinations in a randomized control experiment will allow for a more robust statistical analysis of the performance.

\bibliographystyle{unsrt}  






\end{document}